\pdfoutput=1

\documentclass[11pt]{article}

\usepackage[]{emnlp2022}

\usepackage{times}
\usepackage{latexsym}
\usepackage{graphicx}
\usepackage{amsmath}
\usepackage{cleveref}
\usepackage{bbm}
\usepackage{multirow}
\usepackage{booktabs}
\usepackage{placeins}
\usepackage{enumitem}
\usepackage{todonotes}
\usepackage{color,soul}

\DeclareMathOperator*{\PLM}{PLM}
\DeclareMathOperator*{\VIP}{VIP}
\DeclareMathOperator*{\CQ}{CQ}
\DeclareMathOperator*{\Quantization}{Quantization}
\DeclareMathOperator*{\Contextualization}{Contextualization}
\DeclareMathOperator*{\Prob}{Pr}
\DeclareMathOperator*{\Multinomial}{Multinomial}
\DeclareMathOperator*{\sg}{sg}

\DeclareMathOperator*{\sym}{sim}
\DeclareMathOperator*{\Lce}{L}
\DeclareMathOperator*{\PT}{PT}
\DeclareMathOperator*{\IDPT}{VIP-IDP}
\DeclareMathOperator*{\FT}{FT}

\DeclareMathOperator*{\Adapter}{Adapter}

\DeclareMathOperator*{\T}{T}
\DeclareMathOperator*{\X}{X}
\DeclareMathOperator*{\Y}{Y}
\usepackage{hyperref}

\usepackage{amsmath}
\newcommand\myeq{\mathrel{\overset{\makebox[0pt]{\mbox{\normalfont\tiny\sffamily def}}}{=}}}

\usepackage[T1]{fontenc}

\usepackage[utf8]{inputenc}

\usepackage{microtype}

\title{Vector-Quantized Input-Contextualized Soft Prompts for Natural Language Understanding}

\definecolor{kmy-color}{rgb}{0.858, 0.188, 0.478}

\author{Rishabh Bhardwaj\thanks{\; Equal contribution.}\; $^{\S\ddagger}$\thanks{\; This work was performed while the first author was interning at Salesforce Research. Correspondence to: \;\;<amrita.saha@salesforce.com>} \quad Amrita Saha\footnotemark[1] $\;^{\S}$ \quad Steven C.H. Hoi$\;^{\S}$ \quad Soujanya Poria$\;^{\ddagger}$
\vspace{0.5em}\\
  $^\S$Salesforce Research \\
  $^\ddagger$Singapore University of Technology and Design\\
}

\begin{document}
\maketitle
 \begin{abstract}
Prompt Tuning has been largely successful as a parameter-efficient method of conditioning large-scale pre-trained language models to perform downstream tasks. Thus far, soft prompt tuning learns a fixed set of task-specific continuous vectors, i.e., soft tokens that remain static across the task samples. A fixed prompt, however, may not generalize well to the diverse kinds of inputs the task comprises. In order to address this, we propose \textbf{V}ector-quantized \textbf{I}nput-contextualized \textbf{P}rompts ($\VIP$)\footnote{Our implementation of $\VIP$ is made publicly available  \href{https://github.com/declare-lab/VIP}{https://github.com/declare-lab/VIP}} as an extension to the soft prompt tuning framework. $\VIP$ particularly focuses on two aspects---\textbf{contextual prompts} that learns input-specific contextualization of the soft prompt tokens through a small-scale sentence encoder and \textbf{quantized prompts} that maps the contextualized prompts to a set of learnable codebook vectors through a Vector quantization network. On various language understanding tasks like SuperGLUE, QA, Relation classification, NER and NLI,  $\VIP$ outperforms the soft prompt tuning ($\PT$) baseline by an average margin of 1.19\%. Further, our generalization studies show that $\VIP$ learns more robust prompt representations, surpassing $\PT$ by a margin of 0.6\% - 5.3\% on Out-of-domain QA and NLI tasks respectively, and by 0.75\% on Multi-Task setup over 4 tasks spanning across 12 domains.

\end{abstract}

\section{Introduction}
With the increase in network size, it has become more computationally expensive to fine-tune pretrained language models (PLMs) on downstream tasks. Recent studies have shown that prompt-based learning is a quite effective parameter-efficient method of conditioning the behavior of PLMs for a given predictive task \cite{lester2021power, wu2022idpg}. The goal of prompt-based learning, in contrast to traditional supervised learning approaches, is to find the task-specific template that is augmented to the input text in order to guide the PLM toward generating the correct output \cite{liu2021pre}. The initial efforts in this direction involved manually designed templates to solve natural language processing (NLP) tasks \cite{brown2020language, schick2021few, schick2020exploiting}. 

\begin{figure}[t]
\centering
    \includegraphics[width=0.4\textwidth]{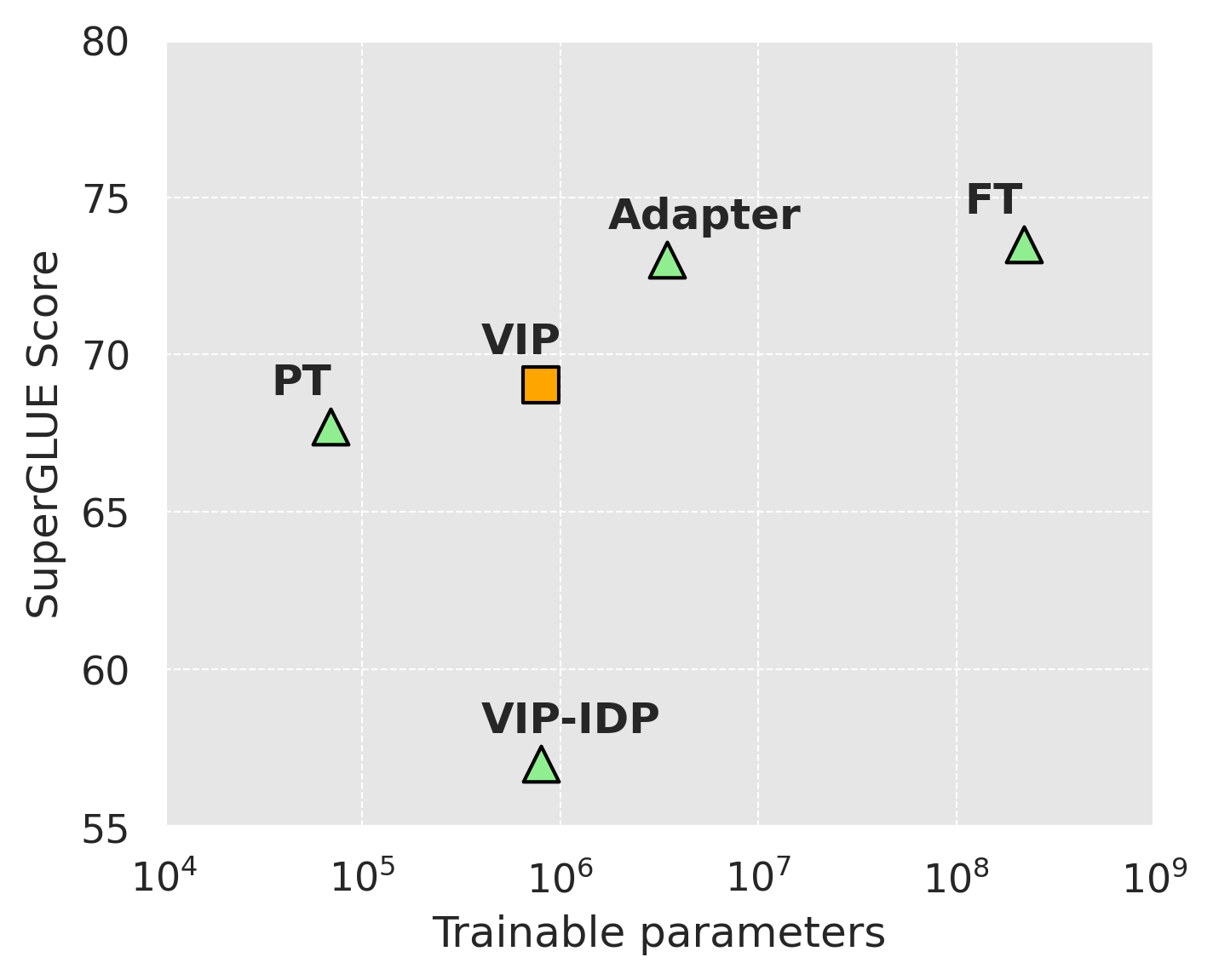}
    \caption{{Comparison of model performance vs parameter size on SuperGLUE dataset. $\PT$ and $\FT$  denotes prompt tuning \shortcite{lester2021power} and fine-tuning respectively while $\Adapter$ \shortcite{houlsby2019parameter} learns adapter modules inserted between layers of the PLM. $\IDPT$ denotes instance dependent prompt generation \shortcite{wu2022idpg} under text-to-text framework. $\VIP$ is our proposed approach of quantized contextual soft prompts. }}
    \label{fig:intro}
    \vspace{-3mm}
\end{figure}

In order to reduce the human effort required in finding appropriate task and LM-specific prompt templates, more recent works aim to automate the process of prompt designing. One of the popular ways is to search for templates in the discrete space of tokens \cite{jiang2020can, haviv2021bertese, yuan2021bartscore}. These are known as hard prompts. Another popular line of research aims at learning prompt token embeddings in a continuous vector space, i.e., soft prompts \cite{li2021prefix, lester2021power}. Our work is an extension of the parameter-efficient soft prompt tuning framework, $\PT$, which we describe below. 

Given a frozen $\PLM$, the soft prompt tuning method $\PT$ \cite{lester2021power} aims to learn a set of soft tokens (vectors) to solve a downstream task. There is a critical limitation of this approach, that is, the learned prompt tokens are expected to perform well across the samples of the task. Thus they are static against the change in input. Henceforth, we refer to them as static tokens. We hypothesize that learning to attend and adapt the prompt tokens to the input provides more flexibility to the system to find the distribution of prompts that generalizes well to diverse forms of the input. To address the limitation of $\PT$, i.e., input-inadaptability of static tokens, we introduce a novel soft prompt tuning technique, \textbf{Vector-quantized Input-contextualized Prompts} ($\VIP$), with two integral properties that are our main contributions:
\noindent
\begin{itemize}[leftmargin=*]
    \item \textbf{Prompt Contextualization.} We propose a ``contextualizer''-- a transformer-based sentence encoder that generates input-adapted prompt tokens from the input text and static soft prompts. While there has been some recent interest in input-dependent prompt generation \cite{idpt,clive2021control,wu2022idpg}, we observe that a straightway utilization of the contextualized tokens as a replacement for static tokens leads to performance degradation (VIP-C in \Cref{tab:qa}). The potential cause is training instability due to noise in the contextual prompt representations that propagates through the frozen $\PLM$ and leads to erroneous predictions. 
    
    \item \textbf{Prompt Quantization.} We tackle the aforementioned challenge by introducing a prompt ``Quantizer'' to reduce the noise in the contextual prompt representations. The quantizer discretizes the continuous space of contextualized prompts, thus allowing us to control its representational capacity. It maps each contextual prompt token to a set of learnable vectors called codebook. These mapped vectors are used as final quantized prompts, with the codebook acting as a parametric discrete latent variable model over the contextual prompt space. Therefore, it is worth noting, quantization limits the space of contextual prompts based on the representation capacity of the codebook vectors. 
    
\end{itemize}

Similar to the other prompting methods, the original input text is concatenated with the obtained quantized (contextual) prompt tokens and subsequently fed into the frozen PLM. The contextualizer and static tokens are learned using backpropagation to minimize the task-specific prediction loss. Whereas the codebook vectors are learned using a more stable update method of exponential moving average \cite{roy2018theory, angelidis2021extractive}.

\begin{figure*}[h]
\centering
    \includegraphics[width=0.8\textwidth]{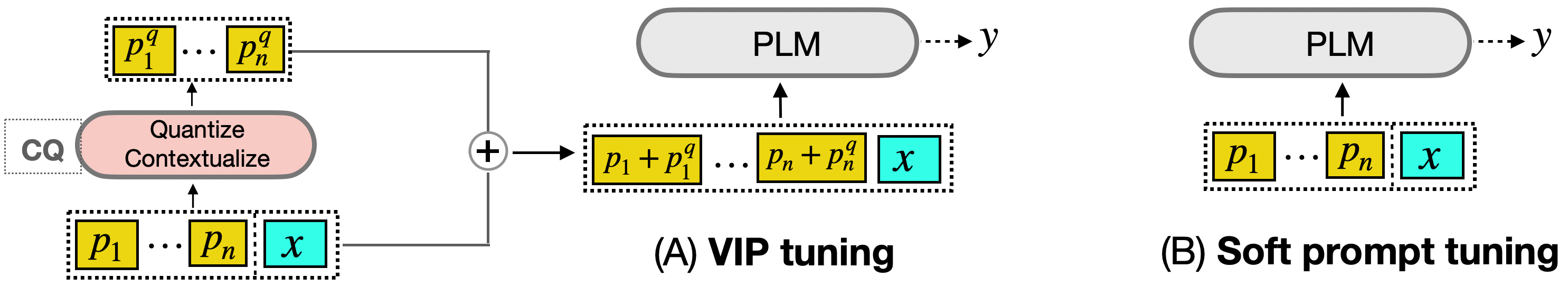}
    \caption{Comparison of VIP with baseline PT. A) \textbf{VIP} tuning denotes the proposed approach where $\CQ$ module contextualizes and quantizes the standard soft prompts. The quantized prompts are then added to the soft prompts through a skip connection. B) \textbf{Soft Prompt tuning} ($\PT$) \cite{lester2021power} is the soft prompt tuning which learns the vectors $p_i$ through backpropagation.
    }
    \label{fig:comp_baselines}
    \vspace{-2mm}
\end{figure*}

Empirically, we demonstrate the utility of our proposed approach VIP on a wide range of NLP tasks. On SuperGLUE, QA, Relation Classification, NER and NLI, VIP improves over the baseline soft prompt tuning method $\PT$ by an average margin of 1.19\%. Additionally we find that VIP outperforms $\PT$ by a margin of 0.6\% - 5.3\% on out-of-domain QA and NLI tasks respectively and by 0.75\% on Multi-Task setup over 4 different tasks. Overall, our experiments showcase that $\VIP$ learns more richer robust prompt representations than static soft prompts $\PT$.

\section{Problem Formulation}
Following \citet{raffel2019exploring}, we cast all the tasks into a text-to-text format. Therefore, tasks such as text and relation classification are reformulated as text generation and the model is expected to generate the class label $y$. Formally, given a pretrained Language Model (PLM) parameterized by $\theta$, we formulate a task $\mathcal{T}$ as conditional text generation $\Prob_{\theta}(\Y | \X)$ where $\X$ and $\Y$ are respectively the sequences of token embeddings corresponding to the texts $x$ and $y$, generated by the tokenizer provided by the frozen PLM.

The prompting approach of \citet{lester2021power} prepends $\X$ with the set of learnable soft tokens $P=\{p_{1}, \ldots, p_{n}\}$. Soft tokens $p_i$ are vectors that lie in a d-dimensional continuous space ${\rm I\!R}^\text{d}$. The idea is to condition the model output $\Y$ directly on $P$ and $\X$ as $\Prob_{\theta}(\Y | P, \X)$, where the static vectors in $P$ are conditionally independent of $\X$ given task $\mathcal{T}$ i.e. $\Prob(P | \X, \T) = \Prob(P | \T)$. $\T$ denotes the set of tokens obtained from task-specific attributes such as task description and class labels. However, such a setting restricts the system to find a singular static set of prompt tokens that is effective for all the diverse samples of a given task.

\paragraph{Motivation.} We hypothesize that to solve a complex language understanding task, the dependence of prompt on input $\X$ is a critical characteristic that aids in generalization over the unseen in and out-of-domain samples. In the parameter-efficient prompt tuning methods such as \citet{lester2021power}, the $\PLM$ is not allowed to tune its parameters and learn task-specific contextualization. This motivates us to learn compact prompt representations in $\VIP$ that can encode task-specific information and contextual information from each of its samples. We represent $\VIP$ tokens by the set of vectors $P^q=\{p_1^q, \ldots, p_n^q\} \in {\rm I\!R}^\text{d}$, which unlike static tokens $P$, holds the desirable property $\Prob(P^q | {\X, T}) \neq \Prob(P^q | {T})$.

\section{Methodology}
In order to obtain input-dependent soft prompts $P^q$, we start with trainable static prompt tokens $P$ similar to \citet{lester2021power}. $P$ along with the original input token sequence $X$ is then fed to a parameterized submodule called $\CQ$, which performs $\Contextualization$ and $\Quantization$. Formally, $P^q = {\CQ}_{\phi}([P, \X])$ where $\phi$ is the parameter set of $\CQ$ module. Our final prompt token representation is obtained by combining the $\CQ$ module's output $P^q$ and the static tokens $P$ through a skip connection i.e. $P+P^q$. This we substitute in place of $P$ in $\Prob_{\theta}(\Y | P, \X)$ from \citet{lester2021power}
\vspace{-0.5em}
\begin{equation} \label{eq:input_to_plm}
    \small
    \Prob(Y|T,P,P^q,\X) \myeq {\PLM}_{\theta}([\T, P+P^q, \X]).
\end{equation}
Next, we elaborate on functioning of the $\CQ$ module, which performs \textbf{Contextualization} followed by \textbf{Quantization} of the soft prompts and is the primary contribution of this work.

\subsection{Contextualization \& Quantization (CQ)}
$\CQ$ expects the input as a sequence of token embeddings, by prepending input $X$ with the input-agnostic soft tokens $P$. As a first step, it performs the token contextualization described below:

\subsubsection{Prompt Contextualization} \label{sec:context}
To reduce the number of trainable parameters and perform meaningful contextualization that is coherent with the input space of $\PLM$, we utilize non-tunable input embedding of $\PLM$ for mapping tokens in $\X$ to respective vectors. Then, as \Cref{fig:cq} shows, $[P, \X]$ is passed through a trainable transformer-based sentence encoder \cite{vaswani2017attention}. For a given task $\mathcal{T}$, $P$ acts as a set of constant vectors. Through the attention mechanism, the encoder fuses $P$ with the context information from $\X$. The output of the sentence encoder is an input-contextualized set of prompt tokens $P^c$. 

\paragraph{Sentence encoder.} To keep the number of trainable parameters low, we perform contextualization in a lower dimensional space. We project $P \in {\rm I\!R}^{n \times d}$ from $d$ to $d'$ dimensions ($d'$<$d$) to obtain $P_l \in {\rm I\!R}^{n \times d'}$. $P_l$ is subsequently fed into a two-layer transformer encoder layers having four attention heads and dense layers of dimension $\{d' \times 2d', 2d' \times d'\}$. The output of this transformer encoder is projected back from $d'$ to $d$-dimension space to get the contextualized prompt token representation $P^c$. The input and output projections are also dense layers with trainable parameters. In our experiments, we have $d$=768 and $d'$=32. The sentence encoder is trained from scratch since it is significantly smaller (~86K parameters) than standard PLMs.

\subsubsection{Prompt Quantization}
The continuous token embeddings $P^c$ at the output of the sentence encoder suffer from high variance in their representations across diverse inputs for a given task, owing to their direct dependence on $X$. This high variance behaves as noise for the frozen PLM, resulting in unstable performance and poor generalization. Another potential limitation of directly utilizing the contextual tokens as prompts is representation collapse resulting in $P^c$ becoming a constant embedding ignoring the input context 
We found that the posterior collapse issue persists unless specifically treated, particularly due to the downstream PLM being frozen.

This inspires us to learn a compact quantized representation $P^q$ of the contextual prompt $P^c$, following \citet{angelidis2021extractive,vandenoord2017vqvae}.
\begin{figure}[t]
\centering
    \includegraphics[width=0.4\textwidth]{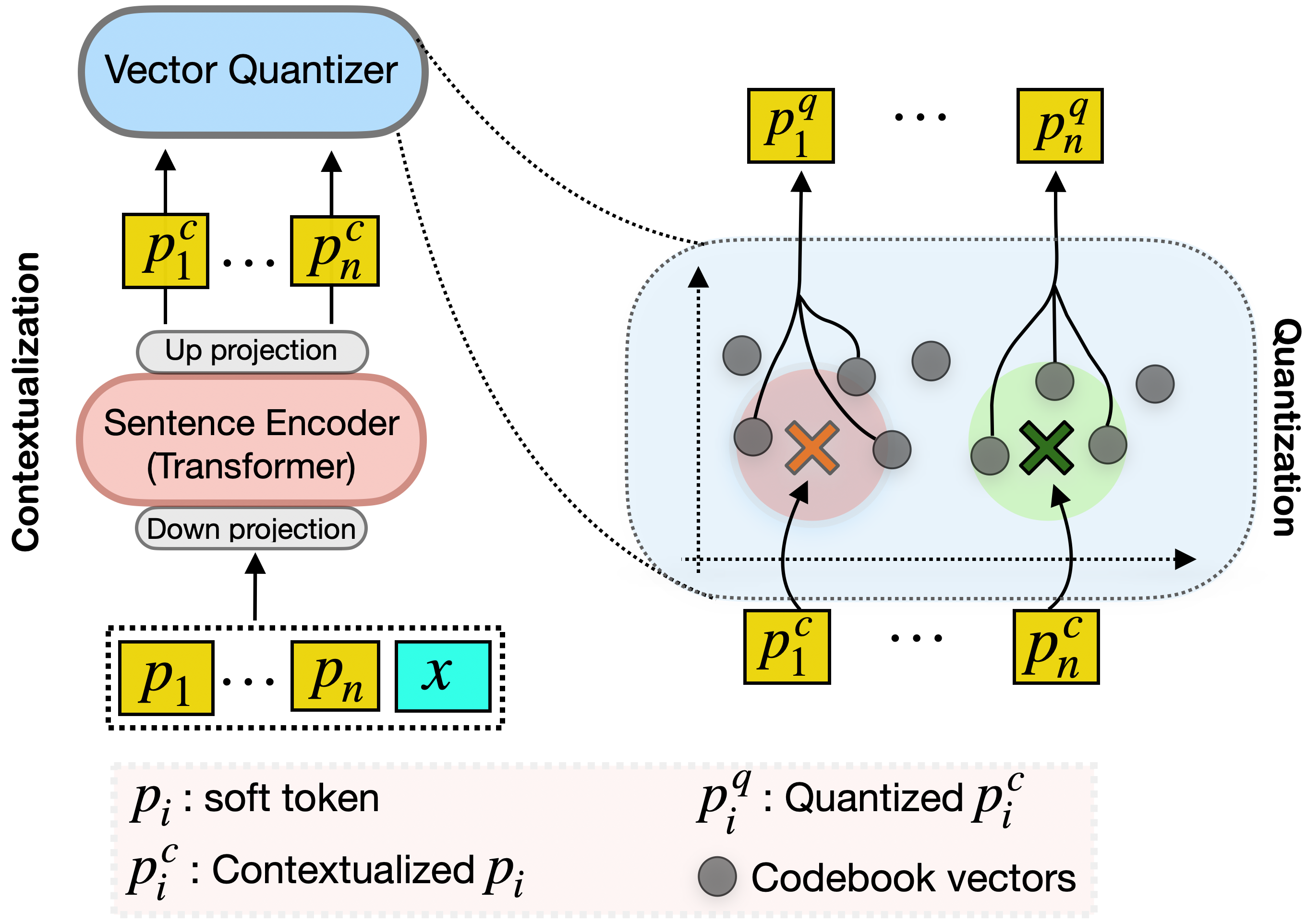}
    \caption{{Contextualizer and quantizer in VIP.}}
    \label{fig:cq}
    \vspace{-2mm}
\end{figure}
Vector quantization achieves a desirable discretization of the latent sentence representations which has been shown to be particularly suitable for language understanding tasks in \cite{roy2018theory, roy2019unsupervised, vandenoord2017vqvae, Mercatali2021DisentanglingGF} owing to the inherent discrete nature of text. 
\paragraph{Vector quantizer.} We maintain a set of $K$ learnable codebook vectors $e \in {\rm I\!R}^{K\times d}$ shared across all the contextualized soft tokens in $P^c$. The quantization is independently performed on each $p_i^c \in P^c$, by sampling a mixture of codebook entries from $e$ that are closest in representation to $p_i^c$ in order to obtain the corresponding quantized prompts $p_i^q$. For this, we first define logits $l_i^k$ for the $i$\textsubscript{th} token $p_i^c$ as negative of its squared euclidean distance from the $k$\textsubscript{th} codebook vector. We then sample, with replacement, $m$ latent codes for each prompt token $p_i^c$, from a Multinomial distribution over the logits $l_i$, which results in the Soft-EM training followed by \citet{angelidis2021extractive,roy2018theory}.
\vspace{-0.5em}
\begin{align}
    \small
    l_i^k &= -\frac{1}{\tau}||p^c_i-e_k||_2^2 \\
    z_i^1, \ldots, z_i^m &\sim \Multinomial (l_i^1, \ldots, l_i^K) \label{Eq:multinomial}
\end{align}
The quantized $p_i^c$, i.e., $p_i^q$ can be computed by averaging over the $m$ samples of codebook entries obtained from the $\Multinomial$ sampling. $p_i^q$'s are discrete latent variables that correspond to centroid of the cluster of the codebook entries closest to $p_i^c$.
\vspace{-0.5em}
\begin{equation}
    \small
        p_i^q = \frac{1}{m}\sum_{j=1}^m e_{z_i^j}
    \vspace{-2mm}
\end{equation}
The contextualization and quantization processes are shown in \Cref{fig:cq}. For experimental stability during the initial phase of training, we scale the logits by a temperature hyperparameter $\tau$ in order to encourage fair codebook usage when sampling using \Cref{Eq:multinomial}. In our experiments $\tau$ is fixed to 100 across all tasks and datasets. It is chosen based on the norm of the codebook vectors in the initial training iterations, which is typically of order $10^2$-$10^3$ for our initialization.

Learning a compact codebook (as described in \ref{subsec:training}) and setting the $\tau$ constant allows the quantizer to explicitly control and bottleneck the information capacity of prompt representations $P^q$ before feeding it to the PLM, thus avoiding posterior collapse. Also, this sampling process ensures that for semantically similar inputs, the quantized prompts will also have similar representations, thus reducing its representation variance w.r.t the input. This consequently helps in stabilizing the frozen PLM's performance, as is evident from our empirical results. Essentially quantization serves as a trade-off between input-agnostic soft prompts and input-contextualized soft prompts, by sampling  prompts from a limited set of learnable codebook vectors.

\paragraph{Skip connection over CQ.} \citet{roy2019unsupervised} illustrated the importance of having a skip connection over the input and output of the quantization bottleneck network. Following this, we also add a skip connection over $\CQ$ module, combining its input and output i.e. $P + P^q$ in order to obtain the final Vector-quantized Input-contextualized Prompt tokens. It is worth noting that the skip connection allows us to effectively fuse the learned information from the input-dependent aspects captured by $P^q$ with the static input-agnostic prompt $P$. Empirically we observe that the information bypass leads to more stable performance i.e. lower variance across random initializations of prompt embedding and sentence encoder parameters.

\subsection{Training}
\label{subsec:training}
The sentence encoder, soft tokens $P$, and codebook vector comprise the set of trainable parameters while we freeze the parameter set of PLM. Our training objective minimizes two losses: i) cross-entropy loss between the output generated by the frozen PLM and the ground truth sequence of tokens and ii) commitment loss which encourages an encoder output to commit to a set of closest codebook vectors, without fluctuating too much. This is important in order to ensure the volume of codebook embedding space does not grow arbitrarily.
\vspace{-0.8em}
\begin{equation} \label{eq:loss_term}
    \small
    \Lce = {\Lce}_{CE} + \beta \sum_{i=1}^{n} ||p_i^c - \sg(p_i^q)||_2^2
\end{equation}
where ${\Lce}_{CE}$ denotes the cross-entropy loss, $\sg$ refers to the stop gradient operator such that $\sg(x)=x$ in the forward pass and zero on the backward pass. $\beta$ is a hyperparameter referred to as the commitment cost, which regulates the extent to which the encoder output should remain close to the sampled codebook vectors.

On back-propagation, the loss term will update the parameters of the sentence encoder and the prompt embedding layer that originally generated $P$. However, a crucial part of the learning involves updating the codebook which we described below.

\paragraph{Learning codebook.} Following works \citet{vandenoord2017vqvae,roy2018theory,angelidis2021extractive} that use codebook-based quantization, we adopt the Exponential Moving Average (EMA) style update of the codebook vectors. This has been shown to result in more stable training than gradient-based updates. 
For each batch, we perform the following two steps:
\noindent
\vspace{-0.5em}
\begin{itemize}[leftmargin=*]
    \item Step 1 counts the number of times j\textsubscript{th} codebook vector is sampled and updates the count $c_j$.
\vspace{-0.5em}
    \begin{equation}
         \small
        c_j = \lambda c_j + (1-\lambda) \sum_{i\in[n], k\in[m] } \mathbbm{1}[z_i^k = e_j]
        \vspace{-2mm}
    \end{equation}
\vspace{-0.5em}
    \item Step 2 updates the embedding of j\textsubscript{th} codebook vector $e_j$  by taking the mean of sentence encoder outputs for which that codebook vector was sampled during $\Multinomial$ sampling.
\vspace{-0.5em}
    \begin{equation}
    \small
    e_j = \lambda e_j + (1-\lambda) \sum_{i\in[n], k\in[m]}\frac{\mathbbm{1}[z_i^k = e_j]p_i^c}{c_j}
    \vspace{-2mm}
    \end{equation}
\end{itemize}
where $\lambda$ is decay parameter set to 0.99, $\mathbbm{1}$ is the indicator function, [n] denotes set $\{1, \ldots, n\}$.

While we choose this online K-means clustering style of quantization of the latent space, there are other discretization techniques e.g. using Gumbel softmax to select the most appropriate codebook vector. We avoid that in our work due to its notorious sensitivity to its temperature hyperparameter. 

\section{Experiments}

\paragraph{Tasks.} We base our experiments on a wide range of language understanding tasks including natural language inference, question answering of extractive and multiple choice nature, relation classification, and named entity recognition.
\vspace{-0.5em}

\begin{table*}[]
\resizebox{\textwidth}{!}{
\begin{tabular}{@{}cccccccccc@{}}
\toprule
\multirow{2}{*}{Model} & CB          & COPA        & WSC         & RTE         & WiC         & BoolQ       & MultiRC     & ReCoRD      & \multirow{2}{*}{Avg} \\ \cmidrule(lr){2-9}
                       & $acc./F1$     & $acc.$        & $acc.$        & $acc.$       & $acc.$        & $acc.$        & $EM/F1_a$      & $EM/F1$      &                      \\ \cmidrule(r){1-1} \cmidrule(l){10-10} 
FT                     & 92.9$_{\pm3.69}$ & 57.0$_{\pm1.0}$    & 63.5$_{\pm0.1}$  & 78.1$_{\pm1.74}$ & 71.9$_{\pm0.7}$  & 79.8$_{\pm0.23}$ & 77.6$_{\pm0.31}$ & 72.3$_{\pm0.32}$ & 74.2$_{\pm0.99}$          \\
Adapter                    & 92.7$_{\pm1.42}$ & 56.7$_{\pm1.53}$ & 65.6$_{\pm3.58}$ & 79.1$_{\pm1.29}$ & 68.9$_{\pm0.57}$ & 79.3$_{\pm0.8}$  & 77.7$_{\pm0.98}$ & 70.2$_{\pm1.76}$ & 73.8$_{\pm1.49}$          \\ \midrule
PT                     & \textbf{75.8$_{\pm3.61}$} & 59.0$_{\pm2.0}$    & 66.7$_{\pm1.48}$ & 70.8$_{\pm2.25}$ & 67.4$_{\pm0.78}$ & 69.3$_{\pm0.6}$  & 69.6$_{\pm0.55}$ & 63.3$_{\pm1.10}$ & 67.7$_{\pm1.55}$          \\
VIP-IDP                & 37.6$_{\pm28.4}$ & 53.0$_{\pm23.7}$   & 66.7$_{\pm1.1}$  & 49.8$_{\pm28.2}$ & 61.2$_{\pm8.16}$ & 62.6$_{\pm0.55}$ & 52.8$_{\pm3.21}$ & 18.2$_{\pm4.91}$ & 57.0$_{\pm12.28}$           \\
VIP-C                  & 73.4$_{\pm2.39}$ & 57.0$_{\pm1.0}$    & 65.1$_{\pm1.43}$ & 70.9$_{\pm1.5}$  & 65.4$_{\pm3.12}$ & 69.2$_{\pm0.71}$ & 69.7$_{\pm0.35}$ & 63.7$_{\pm1.65}$ & 66.7$_{\pm1.52}$          \\
VIP                    & 75.5$_{\pm2.63}$ & \textbf{62.7$_{\pm0.57}$} & \textbf{68.7$_{\pm1.30}$} & \textbf{72.1$_{\pm0.87}$} & \textbf{68.0$_{\pm0.61}$}   & \textbf{69.6$_{\pm1.07}$} & \textbf{70.5$_{\pm0.40}$} & \textbf{65.2$_{\pm0.81}$} & \textbf{69.1$_{\pm1.03}$}       \\ \bottomrule
\end{tabular}}
\caption{Performance of VIP vs. baseline models on SuperGLUE development set. $acc.$, $EM$, $F1$, $F1_a$ and $acc./F1$ denote accuracy, exact match, macro and micro F1 scores and average of accuracy and macro F1 respectively. The numbers in subscript show the standard deviation across 3 random trials using different seeds.}
\label{tab:superglue}
\end{table*}
\noindent
\begin{itemize}[leftmargin=*]
    
    \item \textbf{SuperGLUE} \cite{wang2019superglue}. To gauge general language understanding capabilities of $\VIP$ against baselines, we evaluate on SuperGLUE benchmark using the prescribed evaluation metrics. It comprises of eight tasks---BoolQ, CB, COPA, MultiRC, ReCoRD, RTE, WiC, WSC.
    
\vspace{-0.5em}
    \item \textbf{QA} \cite{fisch2019mrqa}. Here we focus on two types of QA tasks: i) \textit{Extractive QA:} MRQA 2019 shared task consists of multiple question-answering datasets. Six datasets are dedicated each for training and testing while a small portion of training datasets is held-out for model selection. 
    It focuses on evaluating the generalization of a QA system, i.e., how well it performs on out-of-domain examples. ii) \textit{Multiple choice QA:} RACE-middle \cite{race} task to predict the correct answer from four given candidates. Other datasets of this category, MultiRC, and ReCoRD, are incorporated in SuperGLUE.
    
\vspace{-0.5em}
    \item \textbf{Relation classification}. Relation classification aims to identify the semantic relation between two specified entities in a sentence. For this task, we use two datasets---SemEval 2010 task 8 \cite{semeval} and TACRED \cite{tacred}. SemEval is a 19-way classification problem, while TACRED is a larger-scale dataset consisting of 41 relation types.
    
\vspace{-0.5em}
    \item \textbf{NER}. For named entity recognition, we use the English language dataset of CoNLL-2003 shared task \cite{conll}. We frame the same text-to-text format suitable for T5 as in \citet{qin2022lfpt}.
    
\vspace{-0.5em}
    \item \textbf{NLI}. We use Adversarial NLI (ANLI) that is a large-scale benchmark dataset to gauge natural language inference capabilities of the system \cite{nie2019adversarial}. We further evaluate generalization of the model trained on the challenging ANLI task to out-of-domain NLI datasets CB, AXB, AXG, RTE and SNLI. 
    
\end{itemize}

\begin{table}[]
\centering
\resizebox{0.45\textwidth}{!}{
\begin{tabular}{@{}ccccc@{}}
\toprule
\multirow{3}{*}{Model} & \multicolumn{2}{c}{Rel. classification} & NER   & NLI  \\ \cmidrule(l){2-5} 
                       & SemEval             & TACRED            & CoNLL & ANLI \\
                       & $F1_a$                 & $F1_a$               & $F1_a$   & $F1$   \\ \cmidrule(r){1-1}
FT                     & 84.8                & 87.7              & 90.2  & 49.8 \\
Adapter                    & 84.4                & 85.8              & 89.9  & 47.7 \\ \midrule
PT                     & 70.9                & 83.5              & 87.1  & 41.6 \\
VIP-IDP                & 68.95               & 79                & 87.2  & 29.8 \\
VIP-C                  & 71.8                & 83.6              & 87    & 39.3 \\
VIP                    & \textbf{72.4}                & \textbf{84.4}              & \textbf{87.4}  & \textbf{43.2} \\ \bottomrule
\end{tabular}
}
\caption{{Performance comparison on relation classification, NER and NLI tasks.}}
\label{tab:ner_nli_re}
\vspace{-0.3em}
\end{table}

\begin{table*}[]
\centering
\resizebox{0.8\textwidth}{!}{
\begin{tabular}{@{}ccccccccc@{}}
\toprule
Model   & SQuAD & NewsQA & TriviaQA & SearchQA & HotpotQA & Natural QA & RACE-M & Parameter \\ \cmidrule(rl){1-8}
FT      & 83.2  & 65.6   & 80.7     & 77       & 61.6     & 71.1    & 73.6  &~250 M \\
Adapter & 81.4  & 63.9   & 79.3     & 75.1     & 60.1     & 71.6    & 69.9  &2.5 M \\ \cmidrule(rl){1-8}
PT      & 77.5  & 59.3   & 53.4     & 77.6     & 72.9     & 69.6    & 63.4  &0.1 M \\
VIP-IDP & 73.9  & 55.6   & 14.8     & 71.5     & 62.5     & 64.4    & 50.5  &0.9 M \\
VIP-C   & 77.8  & 59.6   & 52.3     & 77.6     & 73.1     & 68.5    & 62.5  &0.1 M \\
VIP     & \textbf{78.7}  & \textbf{61.1}   & \textbf{53.9}     & \textbf{77.9}     & \textbf{73.5}     & \textbf{69.7}    & \textbf{65.9}   &0.9 M \\ \bottomrule
\end{tabular}
}
\caption{{Performance comparison (using F1 score) on Extractive QA from MRQA and multichoice QA RACE-M.}}
\label{tab:qa}
\end{table*}

\begin{table*}[]
\centering
\resizebox{0.8\textwidth}{!}{
\begin{tabular}{@{}cccccccc@{}}
\toprule
\multirow{2}{*}{Train domain} & \multicolumn{7}{c}{Test domain Results in form of F1-Scores of \{PT / VIP\}}                                                                          \\ \cmidrule(l){2-8} 
                              & BioASQ      & DROP        & DuoRC       & RACE        & RE          & TextbookQA  & Micro-Avg.                  \\ \cmidrule(r){1-1}
SQuAD                         & 54.3 / \textbf{54.7} & 29.0 / \textbf{34.4}   & \textbf{35.9} / 35.1 & \textbf{42.6} / 40.4 & 78.7 / \textbf{80.0}   & \textbf{20.8} / 16.6 & 48.9 / \textbf{49.3}          \\
NewsQA                        & \textbf{50.9} / 49.4 & 24.9 / \textbf{25.5} & 36.7 / \textbf{36.9} & 39.4 / \textbf{39.6} & \textbf{73.7} / 73.0   & \textbf{38} / 37.6   & \textbf{48.8} / 48.4 \\
TriviaQA                      & 43.2 / \textbf{46.5} & 17.1 / \textbf{18.1} & 29.4 / \textbf{29.5} & 27.6 / \textbf{29.3} & \textbf{55.5} / 54.1 & 27.7 / \textbf{29.7} & 37.2 / \textbf{37.9}          \\
SearchQA                      & 42.9 / \textbf{48.3} & 19.0 / \textbf{24.3}   & \textbf{27.5} / 26.4 & \textbf{21.9} / 17.8 & \textbf{58.9} / 56.8 & 29.8 / \textbf{32.7} & 38.1 / \textbf{39.2}          \\
HotpotQA                      & \textbf{54.1} / 53.9 & 33.6 / \textbf{34.2}  & 33.7 / \textbf{35.0}  & 36.0 / \textbf{37.2}  & 79.1 / \textbf{80.0}  & 21.1 / \textbf{23.5}  & 49.0 / \textbf{50.0}              \\
NaturalQA                       & \textbf{52.7} / 52.3 & 29.2 / \textbf{31.8} & 29.8 / \textbf{30.5} & \textbf{35.3} / 32.8 & 75.8 / \textbf{77.5} & 27.7 / \textbf{28.3} & 47.4 / \textbf{48.3}          \\ \bottomrule
\end{tabular}}
\caption{{Out-of-domain performance comparison (using F1 Score) of $\VIP$ vs. primary baseline $\PT$ on MRQA.
}} \label{tab:oodQA}
\vspace{-2mm}
\end{table*}

\begin{table}[]
\centering
\resizebox{0.4\textwidth}{!}{
\begin{tabular}{@{}cccccc@{}}
\toprule
\multirow{2}{*}{Model} & CB      & AXB  & AXG  & SNLI & RTE  \\ \cmidrule(l){2-6} 
                        & $acc./F1$ & $acc.$ & $acc.$ & $F1$   & $acc.$ \\ \cmidrule(r){1-1}
PT                      & 43.1    & 63.8 & 49.9 & 62.6 & 70.2 \\
VIP                     & \textbf{59.7}    & \textbf{66.1} & \textbf{51.9} & \textbf{66.7} & \textbf{71.8} \\ \bottomrule
\end{tabular}}
\caption{{ANLI out-of-domain evaluation.}}
\label{tab:anli_ood}
\vspace{-0.2em}
\end{table}

\begin{table*}[h!]
\centering
\resizebox{\textwidth}{!}{
\begin{tabular}{@{}cccccccccccccc@{}}
\toprule
Model      & \multicolumn{1}{l}{ANLI} & \multicolumn{1}{l}{RTE} & \multicolumn{1}{l}{TACRED} & \multicolumn{1}{l}{SemEval} & \multicolumn{1}{l}{ReCoRD} & \multicolumn{1}{l}{RACE-M} & \multicolumn{1}{l}{SQuAD} & \multicolumn{1}{l}{TriviaQA} & \multicolumn{1}{l}{NewsQA} & \multicolumn{1}{l}{SearchQA} & \multicolumn{1}{l}{HotpotQA} & \multicolumn{1}{l}{Nat. QA} & \multicolumn{1}{l}{Avg.} \\ \midrule
PT         & 27.5                     & 60.3                    & \textbf{81.7}                       & 72.6                        & 53.3                       & 57.7                       & \textbf{76.1}                      & \textbf{48.7}                         & 57.0                         & 73.8                         & 69.1                         & 64.7                        & 61.9                     \\
VIP        & 30.6                     & 62.5                    & 80.9                       & \textbf{72.9}                        & \textbf{55.3}                       & 57.9                       & \textbf{76.1}                      & 48.4                         & \textbf{57.9}                       & \textbf{74.2}                         & 69.7                         & 64.8                        & 62.6                     \\ \midrule
VIP: NR    & 38.5                     & \textbf{66.4}                    & 80.8                       & 67.0                          & 53.5                       & \textbf{59.3}                       & 75.1                      & 48.0                           & \textbf{57.9}                       & 73.1                         &\textbf{69.8}                         & \textbf{65.0}                          & \textbf{62.9}                     \\
VIP: NR+DC & \textbf{39.1}                     & 56.3                    & 79.4                       & 64.9                        & 53.0                         & 58.6                       & 75.5                      & 48.2                         & 57.5                       & 73.5                         & \textbf{69.8}                         & 64.6                        & 61.7                     \\ \bottomrule
\end{tabular}}
\caption{{Performance comparison of $\VIP$ and its ablations (from Section \ref{subsec:ablations}) against $\PT$ on multi-task setting.}}
\label{tab:multi_task}
\vspace{-2mm}
\end{table*}

\paragraph{Baselines.} We consider the following baselines. 
\vspace{-0.5em}
\noindent
\begin{itemize}[leftmargin=*]
    \item \textbf{FT} \cite{raffel2019exploring}. It refers to the standard task-specific fine-tuning of PLM.
    
\vspace{-0.5em}
    \item \textbf{Adapter} \cite{houlsby2019parameter}. It inserts learnable modules between layers of a pretrained transformer to perform transfer learning from PLMs to the downstream tasks.

\vspace{-0.5em}
    \item \textbf{PT} \cite{lester2021power}. Soft prompt tuning ($\PT$) is the primary baseline for our work. As shown in \Cref{fig:intro}, $\PT$ prepends a set of continuous vectors $P$ to $\X$ and feeds it to $\PLM$. $P$ is learned via backpropagation through the frozen model. Thus $P$ relies only on the task at hand and choice of PLM while being independent of $\X$.
    
\vspace{-0.5em}
    \item \textbf{VIP-IDP}. Instance-Dependent Prompt Generation Method (IDPG) is a recent approach similar to our idea of making prompt tokens dependent on the input \cite{wu2022idpg}. IDPG cannot be directly treated as a baseline, since it uses RoBERTa-large as PLM and has only restricted applicability to text classification tasks, where it fine-tunes a classifier head over the frozen PLM. Instead, we adopt the IDPG-style prompt generation with our backbone T5-base model and reduce the trainable parameters to be comparable to ours. This version, named $\VIP$-IDP, fetches the input representation from a frozen T5-base encoder and applies two nonlinear projections - first to a low dimension and then upscaling to a space of dimension 768*$n$ to generate $n$ prompt tokens. One limitation is that this network requires two forward passes through the T5-encoder.

\vspace{-0.5em}
    \item \textbf{VIP-C}. This is an ablation of our $\VIP$ architecture, obtained by removing the quantization network. The contextualized prompt tokens from the output of the sentence encoder, $P^c$, are directly augmented to $X$ and fed to the frozen PLM.
    
    \item An important point to note is that while quantization is meaningful for input-dependent prompts, it cannot be applied to standard soft prompts where the prompt tokens have fixed embedding across all input instances of a task.
    
\end{itemize}
\vspace{-0.5em}
\noindent In results \Cref{tab:superglue,tab:ner_nli_re,tab:qa}, bold numbers represent the best performing \emph{prompt}-based model, as $\FT$ and $\Adapter$ enjoy a much larger parameter size. For cross-domain QA, NLI, and multi-task settings we compare $\VIP$ against only primary baseline $\PT$. 

\paragraph{Experimental settings.}
To compare with $\PT$, we base our experiments on the LM-adapted version of T5-base encoder-decoder. We refer to \citet{lester2021power} for finding the best prompt length and fix the prompt token length to $n=100$ for $\PT$, $\VIP$, $\IDPT$. Our quantizer comprises 1000 codebook vectors and $m$ in the multinomial sampling is fixed to 10. We select commitment cost $\beta=0.1$ through grid search over $\{0.01, 0.1, 1\}$. We relegate precise details to \Cref{appendix_exp_setting}.

\subsection{In-domain performance} \label{section: indomain}
In \Cref{tab:superglue}, we report the development set scores on SuperGLUE. Following \citet{lester2021power}, we report results averaging across three random seeds. We find $\FT$ and Adapter with a large number of trainable parameters show significantly better performance than prompt-based methods on the three largest datasets, i.e., ReCoRD, BoolQ, and MultiRC. However, these models perform slightly worse than prompt-based methods on COPA which is a small-scale dataset with only 400 training samples. COPA is a difficult task that requires commonsense causal reasoning, hence reliable tuning a large number of parameters will require a large number of training samples. On the other hand large parameter-sized models can be tuned with less number of samples to solve simpler tasks such as textual entailment in CB.

$\IDPT$'s performance ranks poorest amongst all models across most of the SuperGLUE tasks while also suffering a very high variance across seeds. This is possibly due to straightaway feeding the input-dependent prompts into the frozen PLM. This variance issue is greatly alleviated in $\VIP$ by combining the input-agnostic representation $P$ with the quantized representation $P^q$.

$\VIP$ outperforms the prompt-based baselines on seven out of eight SuperGLUE tasks, while also achieving the lowest variance across all prompt-tuning models. We also notice the drop in performance when the quantization bottleneck is removed ($\VIP$-C). We posit that directly using the sentence encoder's output results in a high variance of the prompt representation w.r.t the input leading to training instability. The performance difference between $\VIP$ and $\VIP$-C is higher for the tasks with lesser training samples such as CB, COPA, and WSC where underfitting is likely, leading to high performance variance.

Likewise in \Cref{tab:ner_nli_re}, we observe that $\VIP$ outperforms all other prompt-based baselines on all three tasks - relation classification, NER and NLI. Particularly on NLI, the improvement is very significant -- 1.5\% better than $\PT$ and 4-14\% better than $\IDPT$ and $\VIP$-C, while on other tasks it improves by 1\% over the prompt baselines.

We next compare our models on question answering. In MRQA shared task, since the test domain is different from the train and development domain, we perform in-domain testing of models on the development data. For model selection, we held out 10\% from the training set. \Cref{tab:qa} compares performance on QA datasets from MRQA task and multi-choice based RACE-M. Across the board, $\VIP$ proves superior to all prompt baselines - outperforming $\PT$ and $\VIP$-C by an average margin of 1\% and $\IDPT$ by over 8\%. However, fine-tuning (FT) and Adapter consistently outperform all prompt models, owing to their large parameter size and due to using T5-base as PLM.

In \Cref{tab:superglue,tab:ner_nli_re,tab:qa}, by comparing VIP with contextualized soft prompt VIP-C and VIP-IDP on 20 datasets across 5 tasks, we empirically substantiate the necessity of a capacity bottleneck.

\subsection{Out of domain performance}
We next evaluate $\VIP$ and its primary baseline $\PT$ on out-of-domain (OOD) MRQA test set. In 5 out of the 6 training domains, the micro-average OOD test performance of $\VIP$ is better than $\PT$, whereas, on NewsQA, both perform comparably. The more fine-grained results show that on all the training domains $\VIP$ performs better than $\PT$ in at least half of the OOD test datasets. Similarly on all the 6 test datasets $\VIP$ consistently outperforms $\PT$ in at least half of the training domains.

Next, in \Cref{tab:anli_ood} we present the cross-domain performance of the ANLI model trained using $\VIP$ and $\PT$ with model selection done on ANLI development set. The test domains consist of the test set of SNLI corpus \cite{bowman-etal-2015-large} and AXB and AXG and the combined train and development set of RTE from SuperGLUE. We observe that $\VIP$ achieves significantly better results on all the OOD NLI tasks as compared to $\PT$. These results empirically justify that the quantized contextual prompts learned by $\VIP$ indeed have a richer and more robust representation than the static soft prompts.

\subsection{Multi-task performance}
\label{subsec:multi-task}
We also perform multi-task experiments considering tasks from NLI (ANLI and RTE), extractive QA (MRQA train set), multi-choice QA (ReCoRD and RACE-M), and relation classification (SemEval and TACRED). For each task, we choose upto a maximum of 10K samples at random for both the training and validation set. For evaluation, we consider the complete standard test split of all the datasets. For SuperGLUE-based tasks RTE and ReCoRD, we consider the test on the validation set. We run the experiments for 50K steps, performing validation at every 2K step with an early stopping set to 5. The validation is done on each dataset separately and the model is chosen based on the best mean validation score. As shown in \Cref{tab:multi_task}, $\VIP$ outperforms its primary baseline $\PT$ in 10 out of 12 tasks and achieves an overall 0.7\% better mean performance. This indicates that our proposed $\CQ$ module is indeed able to learn more robust soft prompts. We posit that a lower score on TACRED and TriviaQA is due to the average validation performance-based model selection strategy.

\subsection{Further Experiments} \label{subsec:ablations}
\paragraph{Dedicated codebook.} In this ablation, instead of having a shared codebook across all prompt tokens, we assign a dedicated codebook matrix (of dimension $\frac{K}{n}$) to each of the $n$ prompt tokens.

\paragraph{Noise resilience.} We also employ noise resilience training of the sentence encoder \cite{gao2021simcse} which outperforms on several tasks as discussed later. We pass the same input to the sentence encoder twice to obtain two sets of representations of the same batch. For a batch with $B$ samples, we define noise as the similarity between two sentence encoder representations ($p^c$), by taking the negative of their Euclidean distance.
With $i$ and $i^+$ representing the two samples obtained by feeding an input instance $i$ twice to the encoder, the sample-wise noise resilience loss $l_i$ is defined to be
\vspace{-0.5em}
\begin{equation}
    \small
    l_i = -\log(\frac{e^{\sym(i,i^+)}}{\sum\limits_{(i \in [B], j \in [B])}e^{\sym(i, j)} + e^{\sym(i, j^+)}}),
\end{equation}
The average sample-wise noise resilience loss over the batch is added to the loss term in \Cref{eq:loss_term}.
\Cref{tab:multi_task} and \Cref{tab:nr_dc} show adding noise resilience and dedicated codebook boosts performance in ANLI, RACE-M, SQuAD, and Nat. QA, while TACRED, TriviaQA and multi-task settings perform better with only noise resilience-based training.
\begin{table}[]
\centering
\resizebox{0.45\textwidth}{!}{
\begin{tabular}{@{}ccccccc@{}}
\toprule
Model      & ANLI & TACRED & RACE-M & SQuAD & TriviaQA & Nat. QA \\ \midrule
PT         & 41.6 & 83.5   & 63.4   & 77.5  & 53.4     & 69.6    \\
VIP        & 43.2 & 84.4   & 65.9   & 78.7  & 53.9     & 69.7    \\ \midrule
VIP:NR     & 43.6 & \textbf{85.0}    & 63.4   & 78.8  & \textbf{54.5}     & 69.9    \\
VIP: NR+DC & \textbf{44.5} & 84.6   & \textbf{66.0}     & \textbf{79.4}  & 54.3     & \textbf{70.2}    \\ \bottomrule
\end{tabular}}
\caption{{Ablations of $\VIP$ with additional Noise Resilience (NR) \& Dedicated Codebook (DC) training.}}
\label{tab:nr_dc}
\vspace{-2mm}
\end{table}
\section{Discussions}
\paragraph{Training-time and memory overhead.} As compared to PT, we did not observe an adverse impact of VIP on training time even with more trainable parameters, and notice a similar convergence pattern. The number of training steps required on SuperGLUE datasets in order (Dataset, PT, VIP) are---(WSC, 180, 90), (CB, 320, 300), (COPA, 377, 442), (WiC, 4080, 1530), (RTE, 546, 760), (BoolQ, 5015, 2360), (RACE-M, 17500, 22000), (MultiRC, 5000, 6000), (ReCoRD, 13500, 19500), where per-iteration time-overhead of VIP is 0.07 seconds (1.96 (VIP) vs 1.89 (PT)). Thus, for WSC, CB, WiC, BoolQ, VIP converges faster. Regarding memory overhead, VIP (97.7MB) is slightly worse (3.3MB) than PT (94.4MB), with the main memory bottleneck being the common frozen T5-base PLM (892MB). While codebook in VIP increases the parameter size, it can be shared across multiple tasks, thus reducing the parameter overhead. In our multi-task experiment, VIP with codebook shared across 3 tasks still outperforms the PT baseline.

\paragraph{Learning codebook for small datasets.} Even on small SuperGLUE datasets such as CB, COPA, and WSC with only 200-400 training instances, VIP performs an average 1.8\% better than PT. Furthermore for CB, the codebook visualization and interpretability analysis (\Cref{codebook_visualization}, \Cref{fig:interp}) shows that VIP indeed learns meaningful representations even for small scale datasets.

\paragraph{Comparison with L2 regularization.} While L2-regularization can also limit the capacity of prompt space, quantization achieves a clustering over the contextualized prompts with the discrete latent variables (codebook vectors) being the clusters. This encourages the quantized prompts to have similar representations for semantically similar inputs, which can neither be achieved in VIP-C nor through L2-regularization. Experiments with L2-regularization did not show any performance gain, in fact, it degraded the performance for a large regularization constant. On SuperGLUE datasets CB and COPA, our pilot-study in the standard PT setting yielded these results in the form (L2-reg constant, Accuracy): For CB-(0,75.8), (1e-4,66.78), (1e-3,70.933), (1e-2,66.77), (1e-1,23.71). For COPA-(0,59.0), (1e-4,53.99), (1e-3,59.20), (1e-2,54.34), (1e-1,45.24). Lastly, learning the codebook can also lead to better interpretability of the prompt space (\Cref{codebook_visualization}, \Cref{fig:interp}).

\paragraph{Comparison with Adapters.} Adapters fall under a different genre of parameter efficient tuning that require amending the PLM to add tunable intermediate layers. This makes them more complicated and expensive than prompt-tuning methods from practical implementation and usage perspective. In general, they are also much less parameter efficient than prompt tuning-based approaches, such as PT and VIP, and hence not directly comparable to them as one of the main baselines.

\paragraph{Comparing by making PT/VIP parameter-size more similar.} Increasing the number of parameters in PT, i.e., prompt length > 100 shows adverse effects on model performance as mentioned by PT paper (Figure3-a of \citet{lester2021power}) and also observed in our experiments. In Appendix \Cref{tab:comb_soft_var}, we investigate different-sized VIP models by varying number of VIP-prompt-tokens and codebook-size, and each of these models still perform better than the best-setting of PT on SuperGLUE.

\paragraph{Impact of prompt quantization.} We observed representation collapse empirically when the model ignores input during the prompt generation step, resulting in prompt representations collapsing to a single fixed embedding. We observed VIP-C (or removing quantization from VIP) is highly prone to this as the contextual prompt representations in the sentence encoder’s output collapse to a constant embedding for several random seeds. VIP's Quantizer alleviates this by limiting the capacity of the prompt space, achieving a clustering over the contextual prompts through the codebook.

\section{Related work}
In this section, we briefly compare with recent prompt-based studies which particularly use dynamic prompts. \citet{clive2021control} proposes a variant of prefix-tuning \cite{li2021prefix} which augments input-dependent activations to each layer, resulting in a much larger parameter size compared to us. Similarly \citet{idpt} also learns a cross-attention network of size 25M over a frozen T5-encoder to generate input-specific prompts, leading to 25x larger parameter size than $\VIP$. \cite{wu2022idpg} is another input-dependent prompt generator that we adapt to our setting as $\VIP$-IDP and compare as a baseline. \citet{jin2022instance} re-scales the soft prompt tokens based on their attention over input tokens, but also fine-tunes the PLM, hence they cannot be considered in a parameter-efficient model tuning regime. 
\section{Conclusion}
We propose a novel extension of soft prompt tuning ($\PT$) - Vector Quantized Input-Contextualized Prompt Tuning (VIP), designed to have two desirable characteristics - (i) contextualizing the soft prompt tokens w.r.t input text using a learnable sentence encoder (ii) discretizing the contextual prompts using a Vector Quantization network. On an extensive set of language understanding tasks - SuperGLUE, QA, NLI, NER, and Relation Classification, $\VIP$ outperforms $\PT$ baseline. Further, our generalization studies on out-of-domain evaluations of QA and NLI and multi-task settings over 4 tasks also show that $\VIP$ is able to learn richer and more robust prompt representations than $\PT$. 

\section{Limitations}
In this section, we point out the limitations of $\VIP$ and its potential future directions.
\noindent
\begin{itemize}[leftmargin=*]
    \item \textbf{Pretraining prompt contextualizer}. The sentence encoder in $\VIP$ is trained from scratch for each downstream task. However, following the prompt pre-training proposed in \citet{gu2021ppt}, a possible future work is to pretrain the prompt contextualizer in a task-agnostic way. 

    \item \textbf{Larger parameter size}. $\VIP$ framework demands a larger parameter size than the baseline soft prompt tuning, owing mainly to the codebook. In Appendix \ref{sec:app_combine_soft_vip} we show that by reducing the number of $\VIP$-prompt tokens and codebook-size, we can reduce the parameter size to one-third while compromising performance slightly on SuperGLUE. More extensive experimental analysis and better techniques for compressing the codebook, we leave as future work.

    \item \textbf{More hyperparameters}. Other than the standard hyperparameters of the sentence encoder, the quantizer introduces new hyperparameters -  codebook-size, multinomial sample size, and the temperature constant $\tau$ to scale logits. While $\VIP$ needs additional hyperparameters, in all our experiments across 20 training datasets from 5 tasks, we fix all hyperparameters related to codebook and sentence-encoder. This shows that our model is indeed not sensitive to the hyperparameters and does not need very specific tuning for each task/setting.

    \item \textbf{Training challenges}. Learning the codebook requires an EMA style updating scheme instead of the standard gradient update. With the PLM being frozen, this needs more careful handling - for e.g. a critical hyperparameter is the value of the temperature constant $\tau$. A very high value can lead to representation collapse of the codebook while very low values can lead to sparse codebook usage. However, as discussed above, $\tau$ is independent of the task and depends on the initial norm of codebook vectors. 

    \item \textbf{Impact on small-scale datasets}. We posit that due to the larger parameter size of $\VIP$, it performs worse than $\PT$ in tasks with lesser training data, e.g. scores on the CB dataset in \Cref{tab:superglue}. This is due to the larger parameter size of $\VIP$. Indeed, by reducing the parameter size of $\VIP$ (in Appendix Table \ref{tab:comb_soft_var}), we achieve much better performance on CB. 

    \item \textbf{T5-base as backbone PLM}. Due to resource limitations, in all our experiments we use T5-base as the backbone. Following \citet{lester2021power} where larger PLMs are shown to improve prompt-tuning performance, we speculate $\VIP$ to showcase a similar effect. Also, though we use T5 as PLM in this work, our $\VIP$ architecture can be used in BERT or GPT style prediction or generation as well. However, a formal analysis of this is left as future work. 
    
    \item \textbf{Data and model bias.} The language understanding tasks and datasets were predominantly in the English language, and thus limit our claims to the English language. Gender, age, race, and other socioeconomic biases may exist in these datasets, and models trained on these datasets may propagate these biases. It is likely that additional biases are also embedded within the T5-base PLM that was used as the backbone of VIP.
\end{itemize}

\section*{Acknowledgement}
Soujanya Poria acknowledges the Ministry of Education, Singapore, under its AcRF Tier-2 grant (Project no. T2MOE2008, and Grantor reference no. MOET2EP20220-0017). Any opinions, findings, conclusions, or recommendations expressed in this material are those of the author(s) and do not reflect the views of the Ministry of Education, Singapore.

\bibliography{acl_latex}
\bibliographystyle{acl_natbib}

\clearpage
\appendix
\section{Appendix}
\label{sec:appendix}

\subsection{Dataset details}
\textbf{SuperGLUE} \cite{wang2019superglue}. To gauge the general language understanding capabilities of $\VIP$ against baselines, we evaluate on SuperGLUE benchmark. It comprises of eight tasks---BoolQ \cite{clark2019boolq}, CB\cite{de2019commitmentbank}, COPA \cite{roemmele2011choice}, MultiRC\cite{khashabi2018looking}, ReCoRD\cite{zhang2018record}, RTE\cite{haim2006second}, WiC \cite{pilehvar2018wic}, and WSC\cite{levesque2012winograd}.

\textbf{MRQA} \cite{fisch2019mrqa}. MRQA comprises of multiple question answering datasets. Six datasets are dedicated each training---SQuAD \cite{squad}, NewaQA \cite{newsqa}, TriviaQA \cite{triviaqa}, SearchQA\cite{searchqa}, HotpotQA\cite{hotpotqa}, NaturalQuestions\cite{naturalquestions}. Six separate datasets are kept for testing---BioASQ\cite{bioasq}, DROP\cite{drop}, DuoRC\cite{duorc}, RACE\cite{race}, RelationExtraction \cite{relationextraction}, TextbookQA \cite{textbookqa}. A small portion of training datasets is held-out for model selection.

\textbf{NLI}. We use Adversarial NLI (ANLI) which is a large-scale benchmark dataset to gauge the natural language inference capabilities of the system \cite{nie2019adversarial}. We further evaluate the generalization of the model trained on the challenging ANLI task to out-of-domain NLI datasets CB, AXB, AXG, RTE (taken from SuperGLUE benchmark) and SNLI \cite{bowman-etal-2015-large}. 

\subsection{Additional Analysis}
\label{sec:app_combine_soft_vip}
\paragraph{\textbf{Combining Soft and VIP Tokens for SuperGLUE}}
\Cref{tab:comb_soft_var} reports SuperGLUE development-set scores for different variants of the $\VIP$ model obtained by taking different combinations of soft prompt tokens and $\VIP$ tokens. We observe that for the configuration with all prompt tokens as $\VIP$, i.e., input-contextualized and quantized, the best average performance is achieved while also having a very low variance across multiple runs. 

Another notable observation is that the performance improves consistently as we increase the number of $\VIP$ tokens replacing static soft prompts. On the other hand, this also shows the trade-off between parameter size and performance Following this analysis, in our main experiments, we choose the configuration of $\VIP$ with 100 $\VIP$-tokens and 0 soft prompt tokens. 

\begin{table*}[]
\centering
\begin{tabular}{@{}cccccc@{}}
\toprule
SuperGLUE & \multicolumn{1}{c}{PT-(100,0)} & \multicolumn{1}{c}{VIP-(80,20)} & \multicolumn{1}{c}{VIP-(50,50)} & \multicolumn{1}{c}{VIP-(10,90)} & VIP-(0,100)               \\ \midrule
\# Tune param        & 76K                 & 315K                & 546k                 &853K                & 930K              \\ \midrule
CB        & 75.83$_{\pm3.61}$                   & 73.43$_{\pm0.50}$                & 78.47$_{\pm3.89}$                & 79.43$_{\pm4.44}$                & 75.47$_{\pm2.63}$              \\
COPA      & 59.00$_{\pm2.00}$                   & 59.33$_{\pm0.58}$                & 58.00$_{\pm2.65}$                & 59.00$_{\pm1.73}$                & 62.67$_{\pm0.57}$              \\
WSC       & 66.67$_{\pm1.48}$                   & 68.26$_{\pm0.95}$                & 65.39$_{\pm0.01}$                & 66.97$_{\pm1.16}$                & 68.73$_{\pm1.30}$              \\
RTE       & 70.77$_{\pm2.25}$                   & 70.17$_{\pm1.46}$                & 72.53$_{\pm1.29}$                & 71.10$_{\pm0.70}$                & 72.10$_{\pm0.87}$              \\
WiC       & 67.37$_{\pm0.78}$                   & 66.50$_{\pm1.37}$                & 66.70$_{\pm1.51}$                & 67.37$_{\pm0.25}$                & 68.00$_{\pm0.61}$              \\
BoolQ     & 69.27$_{\pm0.60}$                   & 70.97$_{\pm0.71}$                & 70.40$_{\pm0.44}$                & 70.01$_{\pm0.38}$                & 69.63$_{\pm1.07}$              \\
MultiRC   & 69.57$_{\pm0.55}$                   & 70.60$_{\pm1.01}$                & 71.53$_{\pm0.29}$                & 72.50$_{\pm0.87}$                & 70.47$_{\pm0.40}$              \\
ReCoRD    & 63.33$_{\pm1.10}$                   & 65.47$_{\pm0.51}$                & 64.37$_{\pm1.00}$                & 64.70$_{\pm1.22}$                & 65.23$_{\pm0.81}$              \\ \midrule
Avg       & \multicolumn{1}{c}{67.72$_{\pm1.55}$}      & \multicolumn{1}{c}{68.09$_{\pm0.88}$}   & \multicolumn{1}{c}{68.42$_{\pm1.38}$}   & \multicolumn{1}{c}{68.89$_{\pm1.34}$}   & \multicolumn{1}{c}{69.04$_{\pm1.03}$} \\ \bottomrule

\end{tabular}
\caption{SuperGLUE Performance of different VIP models obtained by combining $u$ soft prompts and $n$ $\VIP$ prompts denoted by $(u,n)$ and codebook size of 10*$n$. Number of tunable parameters (denoted by \# Tune params) varies across the models due to the different codebook sizes. Numbers in subscript show the standard deviation across 3 randomized trials over different seeds}
\label{tab:comb_soft_var}
\end{table*}

\subsection{Codebook Visualization} \label{codebook_visualization}
In this section we perform visualization and interpretability analysis on the learnt codebook, taking the CommitmentBank (CB) dataset \cite{de2019commitmentbank} as our setting. This is an NLI task where each sample is associated with one of 3 class labels (`neutral', `entailment', `contradiction'). For each instance in the CB development, we first compute the $m$ sampled codebook vectors (i.e. $z_i^1, \ldots, z_i^m$ from Equation  \ref{Eq:multinomial}) that are used to construct each of the 100 $\VIP$ prompt tokens. Thus for each label across all instances in the development set, we can accumulate the information regarding which codebook vectors are sampled for instances of that label. With this, for each label we find the codebook vectors \emph{dedicated} to that label i.e. the codebook vectors that are have high odds of getting assigned to that particular label than to any other label. 

For each soft token, we thus select the three (i.e., number of classes) codebook vectors that are most \emph{dedicated} to each of the 3 class labels. Note that this corresponds to codebook vectors with least-entropy label distribution and the mode of the distribution is that particular class label. 

\begin{figure}[!h]
\centering
    \includegraphics[width=0.5\textwidth]{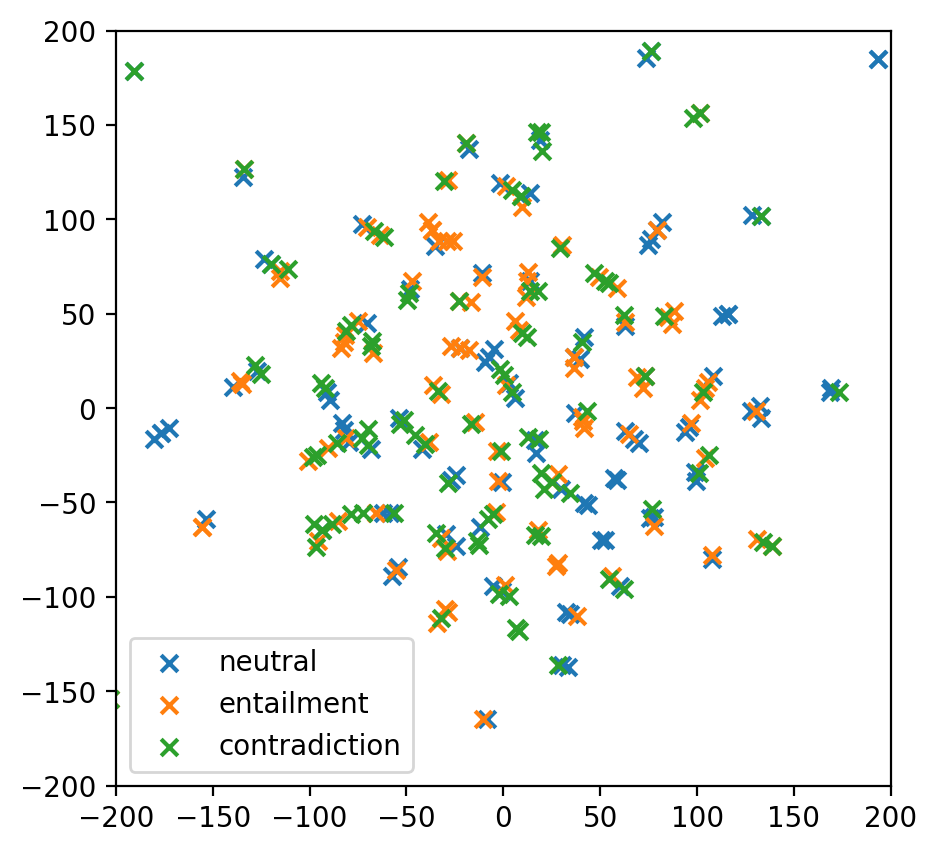}
    \caption{t-SNE map of the learnt codebook representation for NLI task in CB dataset }
    \label{fig:interp}
\end{figure}

\Cref{fig:interp} shows the t-SNE plot of 300 codebook vectors (3 labels $\times$ 100 prompt tokens) selected in this manner. Each codebook vector is color-coded by the corresponding class label it is \emph{dedicated} to. We observe two characteristics of the learned codebooks:
\noindent \begin{enumerate}
    \item codebook vectors of the same color are typically clustered together. These are label-specific codebook vectors that encode exclusive knowledge about that label. This clustering effect is naturally achieved through the EMA style learning of the codebook;
    
    \item there are some clusters with codebook vectors that are dedicated to multiple labels. This we interpret as clusters of codes that capture general label-agnostic information coming from the input context or task description. 
\end{enumerate}

\begin{figure}[!ht]
\centering
\includegraphics[width=0.4\textwidth]{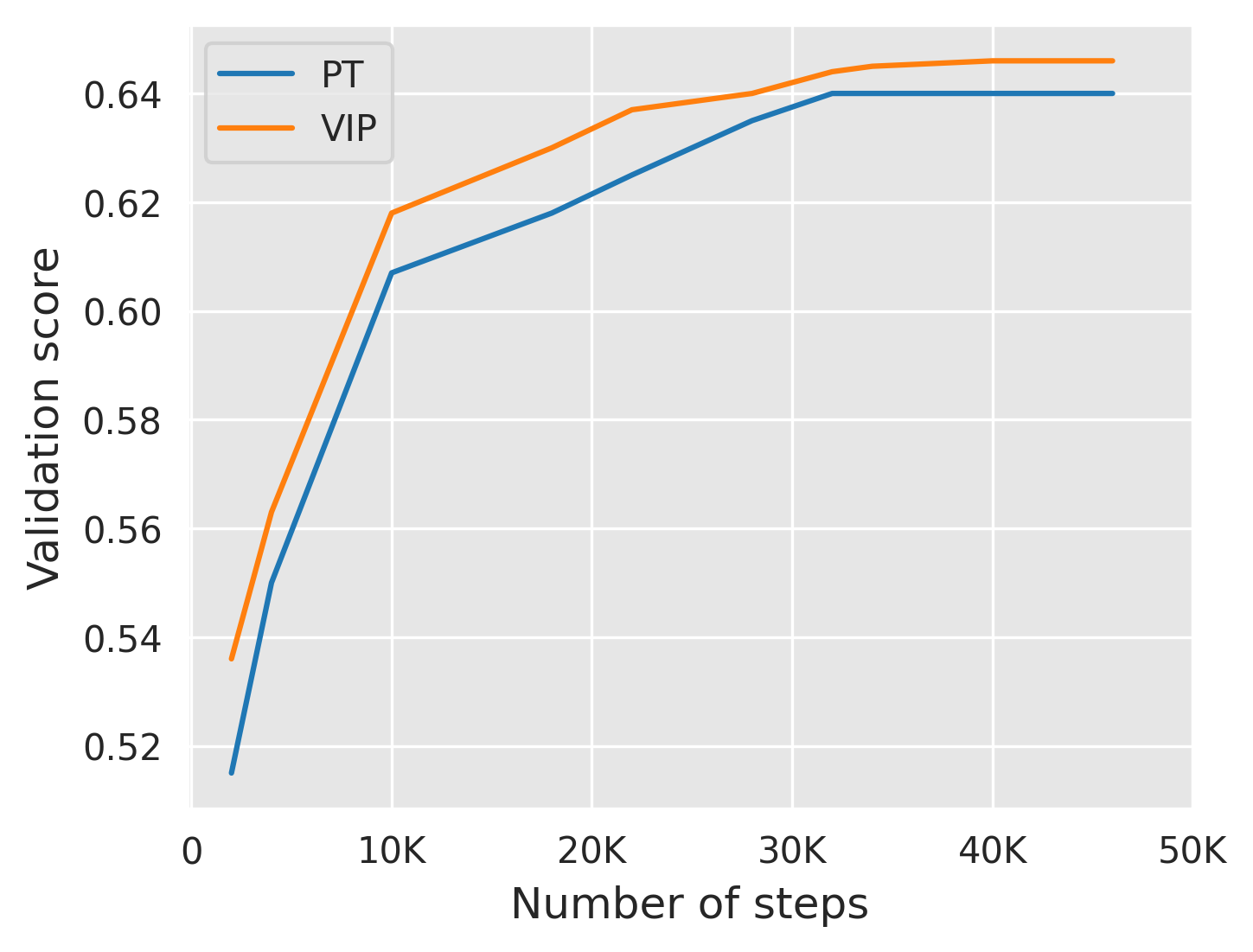}
\caption{Plot of Validation performance on Multi-task dataset vs training iterations. Model selection in Section \ref{subsec:multi-task} and Table \ref{tab:multi_task} is done based on this plot. $\VIP$ achieves more superior performance to baseline $\PT$ throughout all evaluation steps}
\label{fig:multi_task}
\end{figure}

\paragraph{Model Selection in Multi-Task Setting}
Since the multi-task setting constructed by us involves sampling 10K instances randomly for the train and validation sets, we provide additional details on the model selection here. In \Cref{fig:multi_task} we plot the average validation performance of the proposed $\VIP$ model and the baseline $\PT$ method, against the number of training steps, for the multi-task setting reported in Section \ref{subsec:multi-task}. $\VIP$ model is observed to be consistently performing better on the validation dataset than the baseline $\PT$ method at all the evaluation steps. 

\subsection{Experimental Setup} \label{appendix_exp_setting}
\paragraph{Experimental settings.}
For a direct comparison with $\PT$, we base our experiments on the LM-adapted version of T5-base encoder-decoder PLM\footnote{\href{https://huggingface.co/google/t5-base-lm-adapt}{T5-lm-adapt}}. We refer to the study of \citet{lester2021power} to find the best prompt length and fix the number of prompt tokens to $n=100$ for $\PT,\;\VIP$ and $\IDPT$. The sentence encoder settings are previously discussed in \Cref{sec:context}. For the classification tasks such as relation classification and NLI, we prepend the set of label tokens to $[P^q, \X]$ along with the task description (\Cref{eq:input_to_plm}) as non-trainable hard prompt tokens. For $\IDPT$ setting, we keep the weight as learnable parameters in the feedforward generator network. 
This is to enforce the model to rely its predictions on soft tokens which are generated from the input $X$\footnote{experimentally, we found the systems ignore information from $X$ and rely only on the learned bias terms}. $\FT$ feeds $\X$ to the $\PLM$ and fine-tunes the model parameters. Following the same setting as $\FT$, $\Adapter$ learns adapter layer weights as proposed by \cite{houlsby2019parameter} while keeping the $\PLM$ parameters frozen. For all the experiments, standard cross-entropy is used as prediction loss. For optimization, we use Adafactor \cite{shazeer2018adafactor} with constant learning rates (LR), selected via grid search over values $\{0.0001, 0.0005, 0.001\}$ for each of the models, $\CQ$, $\IDPT$, and $\Adapter$. The parameters of the soft prompt ($P$) embedding layer, however, need a significantly different LR, which we set to 0.3, following \citet{lester2021power}. Batch size is set to 128 for the multi-task setup and to 32 for all remaining experiments following \citet{lester2021power}. The quantizer comprises 1000 codebook vectors as parameters learned through EMA. We select commitment cost $\beta=0.1$ through grid search over $\{0.01, 0.1, 1\}$. All our experiments are run for 30K steps, except multi-task setting which is run for 50K steps. We employ different evaluation step sizes for different datasets including epoch-wise evaluation, at 500 steps, and at 2K steps. We tune early stopping according to the evaluation step size and the number of train data samples.

\end{document}